\documentclass[letterpaper]{article} 
\usepackage{aaai24}  
\usepackage{times}  
\usepackage{helvet}  
\usepackage{courier}  
\usepackage[hyphens]{url}  
\usepackage{graphicx} 
\usepackage{natbib}  
\usepackage{caption} 
\usepackage{algorithm}
\usepackage{algorithmic}

\usepackage{newfloat}
\usepackage{listings}
\usepackage{amsmath}
\usepackage{amssymb}
\usepackage{adjustbox}
\usepackage{multirow}
\usepackage{booktabs}
\usepackage{color}
\usepackage{dsfont}
\usepackage{caption}
\usepackage{subcaption}

\usepackage{tikz}
\usetikzlibrary{tikzmark}

\newcommand{\hfilll}{\hspace{0pt plus 1filll}}

\DeclareCaptionStyle{ruled}{labelfont=normalfont,labelsep=colon,strut=off} 
\floatstyle{ruled}
\newfloat{listing}{tb}{lst}{}
\floatname{listing}{Listing}
\title{Weakly Supervised Open-Vocabulary Object Detection}
\author{
    Jianghang Lin\textsuperscript{\rm 1}, Yunhang Shen\textsuperscript{\rm 2}, Bingquan Wang\textsuperscript{\rm 1}, Shaohui Lin\textsuperscript{\rm 3}, Ke Li\textsuperscript{\rm 2}, Liujuan Cao\textsuperscript{\rm 1}\thanks{Corresponding Author}
}
\affiliations{
    \textsuperscript{\rm 1}Key Laboratory of Multimedia Trusted Perception and Efficient Computing, Ministry of Education of China \\ Xiamen University, China. \textsuperscript{\rm 2}Tencent Youtu Lab, China.
    \textsuperscript{\rm 3}School of Computer Science and Technology \\ East China Normal University, Shanghai, China.\\
    \{hunterjlin007, wangbingquan\}@stu.xmu.edu.cn, \\ \{shenyunhang01, shaohuilin007, tristanli.sh\}@gmail.com, caoliujuan@xmu.edu.cn, 
}

\usepackage{bibentry}

\begin{document}

\maketitle

\begin{abstract}
Despite weakly supervised object detection~(WSOD) being a promising step toward evading strong instance-level annotations, its capability is confined to closed-set categories within a single training dataset.
In this paper, we propose a novel weakly supervised open-vocabulary object detection framework, namely WSOVOD\footnote{Source code are available at: \color{purple}{https://github.com/HunterJ-Lin/WSOVOD}}, to extend traditional WSOD to detect novel concepts and utilize diverse datasets with only image-level annotations.
To achieve this, we explore three vital strategies, including dataset-level feature adaptation, image-level salient object localization, and region-level vision-language alignment.
First, we perform data-aware feature extraction to produce an input-conditional coefficient, which is leveraged into dataset attribute prototypes to identify dataset bias and help achieve cross-dataset generalization.
Second, a customized location-oriented weakly supervised region proposal network is proposed to utilize high-level semantic layouts from the category-agnostic segment anything model to distinguish object boundaries.
Lastly, we introduce a proposal-concept synchronized multiple-instance network, \textit{i.e.}, object mining and refinement with visual-semantic alignment, to discover objects matched to the text embeddings of concepts.
Extensive experiments on Pascal VOC and MS COCO demonstrate that the proposed WSOVOD achieves new state-of-the-art compared with previous WSOD methods in both close-set object localization and detection tasks. Meanwhile, WSOVOD enables cross-dataset and open-vocabulary learning to achieve on-par or even better performance than well-established fully-supervised open-vocabulary object detection~(FSOVOD).
\end{abstract}

\begin{figure}[t]
     \centering
     \begin{subfigure}[b]{0.42\textwidth}
         \centering
         \includegraphics[width=0.99\textwidth]{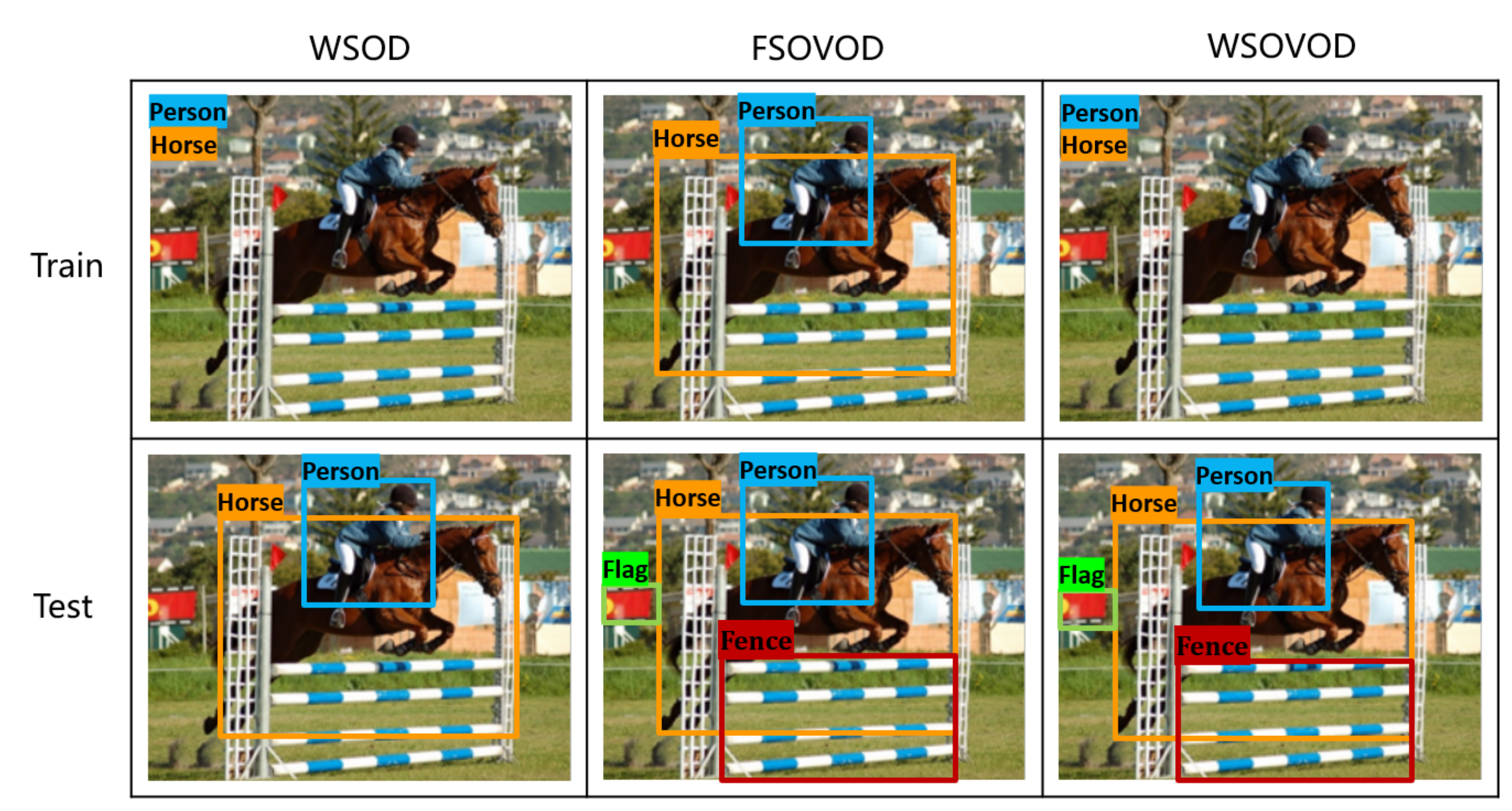}
         \caption{Paradigm comparisons.}
     \end{subfigure}
     \begin{subfigure}[b]{0.45\textwidth}
         \centering
         \includegraphics[width=0.99\textwidth]{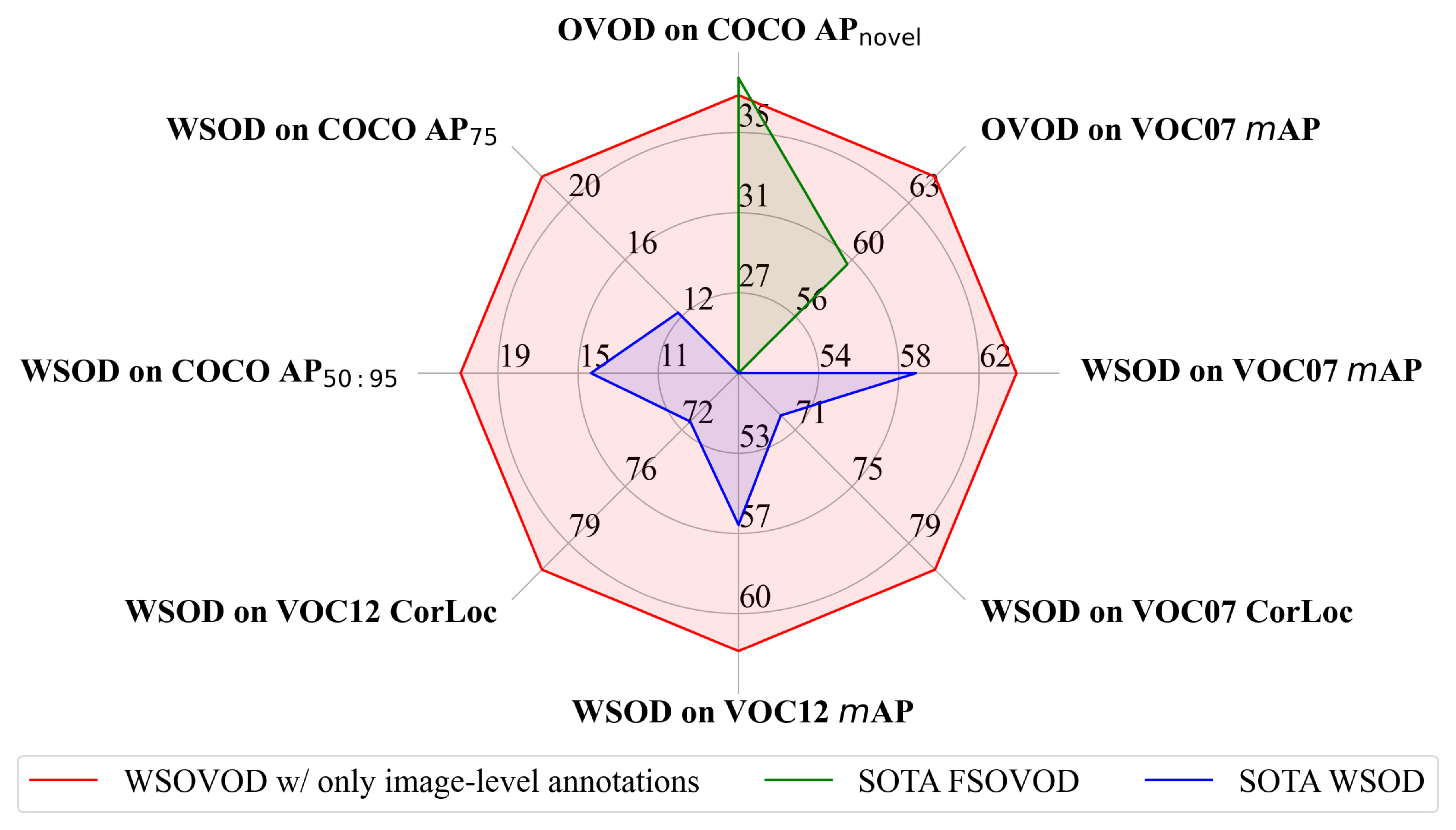}
         \caption{Qualitative comparisons.}
     \end{subfigure}
\caption{
Vanilla WSOD is confined to detecting known categories from the training set, \emph{e.g.}, \textit{Person}, and \textit{Horse}.
The proposed WSOVOD is generalized to unseen categories, \emph{e.g.}, \textit{Flag}, and \textit{Fence}.
WSOVOD outperforms the previous state-of-the-art WSOD methods and achieves on-par or even better performance than FSOVOD.
}
\label{fig_comparison}
\end{figure}

\section{Introduction}
\label{sec:intro}
In the past decades, the artificial intelligence community has witnessed great progress in object detection.
%
In particular, large amount of human-annotated datasets significantly promotes the prosperity and progress of fully-supervised object detection~(FSOD), such as Faster RCNN~\cite{NIPS2015_14bfa6bb}, YOLO~\cite{Redmon_2016_CVPR,8100173}, Detr~\cite{carion2020end} and their variants~\cite{zhu2020deformable,zheng2020end,sun2021rethinking}.
Nevertheless, the laborious and lavish collection of instance-level annotations has severely barricaded the applicability of FSOD in practical application with large-scale categories. 
By excavating image-level category supervision that indicates the presence or absence of an object, weakly supervised object detection~(WSOD) has attracted much attention recently since image-level annotations are widely available in easily-collected classification-like datasets.
%

Unfortunately, a \textit{de facto} limitation of the existing WSOD methods~\cite{tang2017multiple,tang2018pcl,kim2020tell} stems from their detectors only concentrating on few categories concepts within individual datasets, such as $20$-category Pascal VOC~\cite{everingham2010Pascal} and $80$-category MS COCO~\cite{lin2014microsoft}.
Little effort has been made to explore the limit of WSOD learning at scale toward detecting novel objects.
Thus, it may not fully exploit the latent capacity of WSOD whose original intention is to leverage the tremendous amount of tagged images to train object detectors.
To solve the above limitation, as illustrated in Fig.\,\ref{fig_comparison} (a), we extend WSOD settings to detect and localize open-vocabulary concepts using joint large-scale weakly-annotated datasets that are publicly available.
Accordingly, a weakly supervised open-vocabulary object detection, referred to as WSOVOD is put forth.
To this end, three main challenges, as we start in this paper, obstacle to the implementation of WSOVOD. 
First, non-identical data distributions may bring dataset bias~\cite{Kim2021,Torralba2011,Jiang2022} to affect the feature learning, hindering the vision-language alignment introduced as followed.
For example, ILSVRC~\cite{ILSVRC} is an object-centric dataset with a balanced category distribution, while LVIS~\cite{Gupta2019} has many complex scenes with Zipfan distribution.
Second, the reliance of existing WSOD methods upon traditional object proposal generators prevents models from learning proposal extraction at different semantic levels, since they only use low-level features computed on super-pixel~\cite{Felzenszwalb2004} or counters~\cite{Dollar2013}.
Though UWSOD~\cite{shen2020uwsod} learns object proposal generator under pseudo-label supervision, it is difficult to generate high-quality proposals due to noisy training.
Third, weak supervision hardly aligns vision-language representation.
%
In the existing open-vocabulary studies~\cite{Ma2022,gu2021open,Zang2022}, the visual-semantic alignments are realized in a fully-supervised manner where classification embeddings and box knowledge are necessary.
Though recent methods~\cite{zhou2022detecting,Kamath,zareian2021open} resort to weak information, \textit{e.g.}, captions, they deeply rely on strong box-level annotations.
To solve the above three problems and overcome the limitations of common WSOD approaches, our WSOVOD framework (in Fig.\,\ref{fig_wsovod_framework}) innovates in three aspects:
1) We extract data-aware features to generate for each image input-conditional coefficients and combine dataset attribute prototypes to identify dataset bias in proposal features of different distributions.
Explicitly, an additional branch learns to squeeze the global image feature into a channel-wise global vector as coefficients to weight dataset attribute prototypes for re-calibrating final proposal features, which effectively helps achieve model generalization to different scenarios and categories.
2) A location-oriented region proposal network is proposed to leverage high-level semantic layouts from the image segmenter to distinguish object boundaries.
Recent interactive segmentation work SAM~\cite{kirillov2023segany} exhibits strong image segmentation capabilities, but it lacks semantic recognition ability.
Here, we transfer the knowledge from SAM to a customized region proposal network upon high-quality proposals.
3) We introduce a proposal-concept synchronized multiple-instance network that implements object mining to discover objects under image-level classification embeddings, as well as instance refinement to align vision-language representation.
Specifically, we obtain text embeddings of the target vocabulary from the pre-trained text encoder, which are considered as category prototypes for multiple-instance learning.
Also, we transform the multi-branch refinement heads in the common WSOD framework into open-vocabulary learning to further align object and concept representation.
In addition, we leverage SAM to refine the box coordinates of the supervision between multi-branch refinement heads.
Extensive experiments on Pascal VOC and MS COCO demonstrate that the proposed WSOVOD achieves on-par or even better performance compared to fully-supervised open-vocabulary detection methods, which paves a new way to explore the large number of visual concepts from image-level supervisions.
For example, our method significantly outperforms OVR-CNN~\cite{zareian2021open}, ViLD~\cite{gu2021open} and Detic~\cite{zhou2022detecting} that require box-level annotations of base categories, by $13.9\%$, $9.1\%$ and $8.9\%$ AP, respectively, for novel categories in MS COCO.
Moreover, WSOVOD achieves new state-of-the-art performance compared to the previous WSOD methods under the close-set and single-dataset settings while being able to detect novel categories.
\section{Related Work}
\label{sec:related_work}
\subsection{Weakly Supervised Object Detection}
\label{sec:rw_wsod}
Combining multiple-instance learning~(MIL)~\cite{dietterich1997solving} with convolutional neural networks~(CNNs) has made great progress in WSOD.
WSDDN~\cite{bilen2016weakly} is the prior work to introduce MIL into CNN and model WSOD as a proposal classification.
However, WSDDN suffers from local optimization problems since the detector tends to detect high-activated regions.
OICR~\cite{tang2017multiple} further attaches multi-branch refinement to WSDDN, which gradually propagates the scores of the salient regions to the complete objects.
These methods are highly dependent on traditional proposal generation methods~\cite{uijlings2013selective,pont2016multiscale} 
and do not regress the final proposal boxes.
Furthermore, UWSOD~\cite{shen2020uwsod} learns multi-scale features and the region proposal network in an end-to-end unified framework.
Nevertheless, the region proposal network is prone to be saturated due to the noisy pseudo-ground-truth boxes in the early training period, which has inferior performance than the cutting-edge WSOD methods.
Different from these methods, we exploit knowledge transfer from the category-agnostic segmenter to pursue high-quality and high-recall object proposals.
\begin{figure*}[t]
\centering
\includegraphics[width=0.90\textwidth]{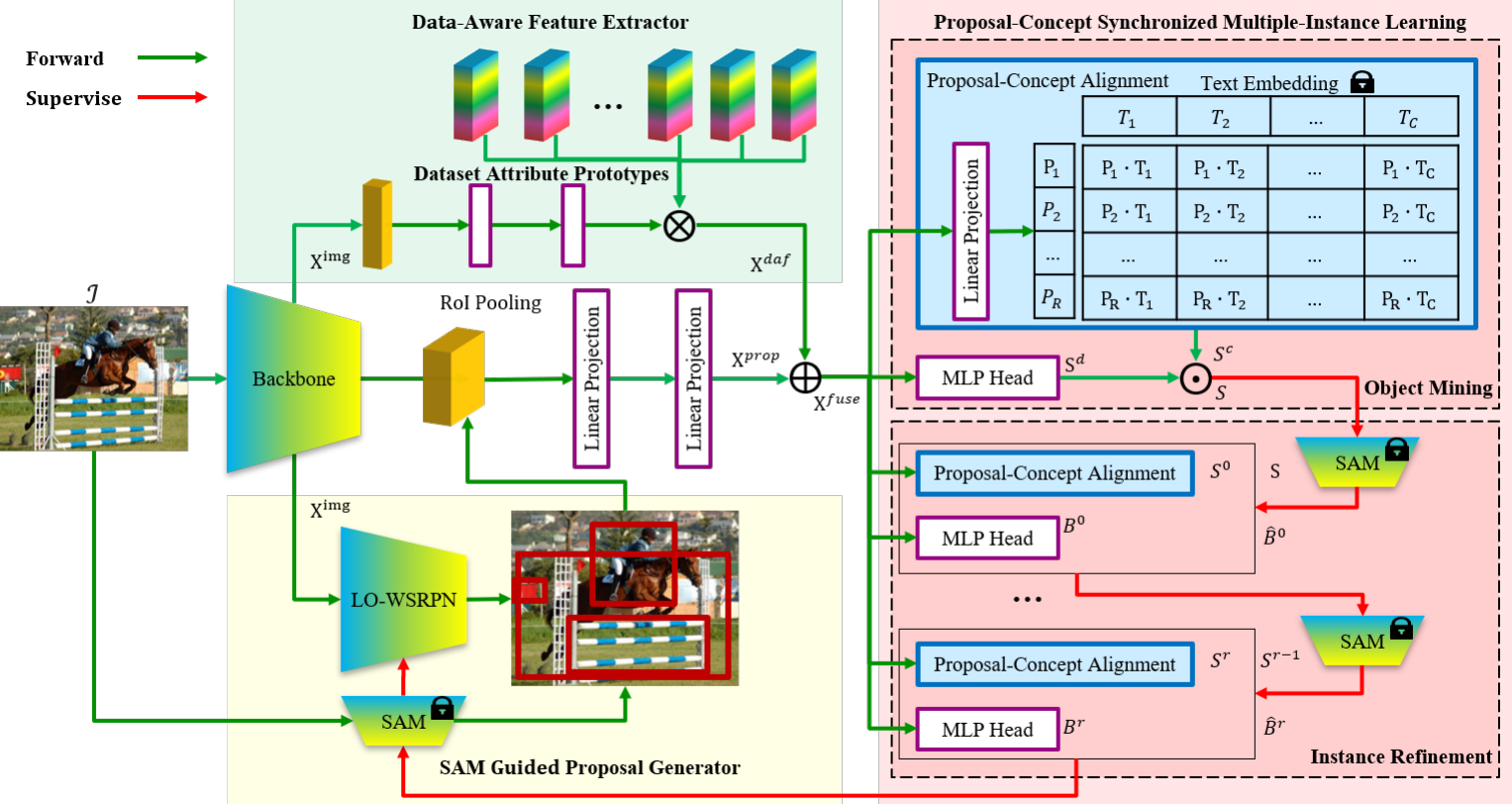}
\caption{Illustration of the proposed WSOVOD framework.
The proposal generator combines candidate regions from LO-WSRPN and SAM that may potentially contain objects for subsequent object mining.
The data-aware feature extractor outputs unbiased dataset attribute features by identifying dataset bias from dataset attribute prototypes.
The proposal-concept synchronized multiple-instance learning discovers potential objects that match the target vocabularies in image-level labels.
}
\label{fig_wsovod_framework}
\end{figure*}
\subsection{Open-Vocabulary Object Detection}
\label{sec:rw_ovod}
Open-vocabulary object detection~(OVOD)~\cite{zareian2021open,gu2021open,minderer2022simple,zhou2022detecting} is an attractive research topic in recent years, whose goal is to detect unseen or novel classes that occupy a particular semantically-coherent region within an image. 
OVOD differs from zero-shot object detection~\cite{bansal2018zero} in that it can access large-scale novel objects with weakly-supervised labels, \textit{e.g.}, tags, and captions.
However, they share the same paradigm of learning a cross-modal vision-language representation space to model image regions and word descriptors.
The main challenge in this field is aligning proposal features with category text embeddings, thus it is crucial to use image-text knowledge efficiently~\cite{radford2021learning,li2022grounded}.
OVR-CNN~\cite{zareian2021open} pre-trains the detector on image-text pairs using contrastive learning and fine-tunes it on detection data with a limited vocabulary.
OWL~\cite{minderer2022simple} further transfers knowledge from vision-language models to transformer-based detectors with contrastive image-text pre-training and detection fine-tuning.
Detic~\cite{zhou2022detecting} improves OVOD performance of long-tail categories via image-level annotated data, such as ImageNet~\cite{deng2009large} and conceptual captions~\cite{sharma2018conceptual}.
Different from these approaches, our proposed WSOVOD uses MIL-based object mining to discover potential objects and refines them by multi-branch refinement open-vocabulary heads gradually.
All of the above methods are highly dependent on bounding-box annotations, while WSOVOD is devoted to efficiently exploring weakly-annotated data.
\section{Methodology}
\label{sec:method}
As illustrated in Fig.\,\ref{fig_wsovod_framework}, an image $\mathcal{I}$ first goes through the vision backbone to extract global image features $X^\mathrm{img}$.
Then, the data-aware feature extractor takes in $X^\mathrm{img}$ to generate coefficients for combining dataset attribute prototypes as data-aware features $X^\mathrm{daf}$.
Meanwhile, a proposal generator also takes in $X^\mathrm{img}$ to hypothesize object locations.
Next, RoI pooling 
crops the pooled features from global feature $X^\mathrm{img}$, and two fully-connected layers transform them to get proposal features $X^\mathrm{prop}\in \mathcal{R}^{R \times D}$, where $R$ is proposal number in image $\mathcal{I}$ and $D$ is feature vector length.
We further fuse proposal features $X^\mathrm{prop}$ with data-aware features $X^\mathrm{daf}$ to deal with dataset bias, resulting in $X^\mathrm{fuse}$.
Finally, 
a proposal-concept synchronized multiple-instance learning takes in $X^\mathrm{fuse}$ to discover objects constrained by image-level classification embeddings and align representation between objects and concepts.
The overall training objective function is formulated as:
\begin{equation}
    \mathcal{L}_\mathrm{WSOVOD} = \mathcal{L}_\mathrm{PG} + \mathcal{L}_\mathrm{OM} + \mathcal{L}_\mathrm{IR},
    \label{eq:total_loss}
\end{equation}
where $\mathcal{L}_\mathrm{PG}$ is proposal generator loss.
And $\mathcal{L}_\mathrm{OM}$ and $\mathcal{L}_\mathrm{IR}$ are object mining and instance refinement losses for proposal-concept synchronized multiple-instance learning.
\subsection{Data-Aware Feature Extractor}
\label{sec:method_dafe}
To better align vision-language representation, it is necessary to learn as many categories as possible, however, an individual dataset contains limited concepts.
This motivates us to train one detector upon multiple datasets jointly to generalize the detection scope of WSOD.
The main challenge stems from domain incompatibility over non-identical data distribution. 
Much of the bias can be accounted for by the divergent goals of the different datasets: 
For example, LVIS~\cite{Gupta2019} has an average of $11.2$ instances from $3.4$ categories per image with long-tail Zipfan distribution, while most images in ILSVRC~\cite{ILSVRC} are object-centric with single category.
Such variant dataset biases hurt the representation learning, thus simply combining multiple datasets has poor performance as observed in our experiments.
In contrast, such bias could be well recognized even from a single image by humans and classifiers~\cite{Torralba2011}.
To this end, we design a data-aware feature extractor~(DAFE) to generate generalized dataset-level features for cross-dataset learning with different scenarios and different categories.
The key intuition of DAFE is to capture the unique and identifiable ``signature'' of each dataset conditioned on full-image information and adjust proposal features accordingly.
From an alternative perspective, DAFE shifts the proposal position in the visual-semantic embedding space via visual prompt, as text embeddings are fixed.
Specifically, it consists of a global average pooling layer to squeeze the input information from image feature maps $X^\mathrm{img}$.
Then two fully-connected layers followed by the Tanh activation function learn to generate coefficients based on the image input to combine dataset attribute prototypes for identifying the dataset bias from the squeezed global features and produce data-aware feature $X^\mathrm{daf}$ with the same dimension of the proposal features.
Finally, we aggregate $X^\mathrm{daf}$ with the proposal features $X^\mathrm{prop}$ by element-wise summation: $X^\mathrm{fuse} = X^\mathrm{prop} + X^\mathrm{daf}$.
Thus, the input-conditional vector $X^\mathrm{daf}$ aims to re-calibrate the original proposal features to de-bias the different distributions, which are then used for the subsequent open-vocabulary object mining and refinement. 
Albeit the apparent simplicity, DAFE approach plays a vital role in cross-dataset learning.
\textbf{Discussion.} The proposed DAFE, to some extent, is related to recent prompt tuning~\cite{Jia2022} 
that adapts large foundation models to downstream tasks with a small amount of task-specific learnable parameters.
Our approach differs from theirs in two folds.
First, most prompt learning methods perform data-space adaptation by transforming the input.
For example, approaches in~\cite{Feng2022}  
append a sequence of learnable vectors to the textual input, and method in~\cite{Bahng2022} learns an image perturbation to convert the image to the formats of downstream tasks.
Different from the above methods, our DAFE eliminates the different dataset distributions by feature-space adaptation with an input-conditional vector.
Second, existing prompt tuning mainly focuses on fully-supervised learning, which is difficult to generalize to wide unseen categories.
Input-conditional prompt learning~\cite{Zhou2022a} still relies on an online text encoder to generate input-specific weights for each image.
Our adaptation does not require an online text encoder during training and inference.

\subsection{SAM Guided Proposal Generator}
\label{sec:method_samgpg}
Most existing WSOD methods use traditional proposal methods with low-level features to generate region candidates, which prevents the models from end-to-end learning with high-level semantic information.
We design a location-oriented weakly supervised region proposal network~(LO-WSRPN) to recognize category-agnostic potential objects, which further transfers knowledge of high-level semantic layouts from SAM~\cite{kirillov2023segany}.
In detail, similar to RPN from Faster RCNN~\cite{NIPS2015_14bfa6bb}, LO-WSRPN has a $3\times3$ convolutional layer with $256$ channels followed by three sibling $1\times1$ convolutional layers for localization and shape quality estimations, respectively.
The first two convolutional layers are responsible for localization quality estimation, predicting centerness $c$ and foreground probabilities $p$, respectively.
We use $s=\sqrt{c \cdot p}$ as the localization quality for each region proposal during inference.
The last convolutional layer is responsible for shape quality estimation.
Different from anchor-based detectors, we directly view locations as training samples instead of anchor boxes.
We replace the standard box-delta targets $(x,y,h,w)$ with distances $t = (l,r,t,b)$ from the location to four sides of the ground-truth box as in~\cite{tian2019fcos}.
The training objective function of this module is formulated as:
\begin{equation}
    \begin{split}
        \mathcal{L}_\mathrm{PG} = 
        & \mathcal{L}_\mathrm{BCE}(p,p^*) + \\
        & \mathds{1}_{p^*=1} \{L_1(c,c^*) 
        + \mathcal{L}_\mathrm{IoU}(t,t^*)\} ,
    \end{split}
    \label{eq:object_proposals_generator_loss}
\end{equation}
where $\mathcal{L}_\mathrm{BCE}$ is the binary cross-entropy loss function, $L_1$ constrains the distance between the sampling anchor points and the pseudo-ground-truth~(PGT) boxes, $\mathcal{L}_\mathrm{IoU}$ measures the shape difference between the predicted boxes and the PGT boxes, thereby constraining the shape of the predicted boxes.
We use WSOVOD's final predictions as PGT boxes and assign the corresponding targets, \textit{i.e.}, $p^*$, $c^*$, and $t^*$.
However, object proposals from  LO-WSRPN are extremely noisy in the early stage of training as observed in~\cite{shen2020uwsod}, which has a negative impact on subsequent object mining and brings in inferior PGT boxes.
Inspired by large-scale interactive segmentation models~\cite{kirillov2023segany}, we leverage SAM to generate additional proposals during training, which helps stabilize subsequent object mining.
In detail, we first sprinkle evenly $32\times32$ grid points as the prompt input of SAM to generate additional proposals.
Then, we concatenate the SAM proposals with the learned proposals from LO-WSRPN as input to subsequent object mining.
Incorporating knowledge from SAM not only helps enrich the high-quality object proposals but also leverages high-level semantic layouts from the image segmenter to distinguish object boundaries.

\subsection{Proposal-Concept Synchronized Multiple-Instance Network}
\label{sec:method_om}
The central idea of fully-supervised open-vocabulary object detection~(FSOVOD) is to align object features with text embeddings which are pre-trained on large-scale image-text pairs like CLIP~\cite{radford2021learning}.
In detail, FSOVOD methods 
convert a generic two-stage object detector to an open-vocabulary detector by replacing the learnable classifier head with fixed language embeddings, corresponding to the category names.
Thus, object-level annotations are required during training to maximize the embedding similarities of positive region-category pairs and minimize that of negative ones.
However, it is challenging to align object-level vision-language representation with only image-level supervision.
Fortunately, WSOD is often formulated as multiple-instance learning~(MIL)~\cite{dietterich1997solving} and implicitly learns instance-based classifier from image-level information.
Therefore, our WSOVOD extends the common MIL-based WSOD framework~\cite{bilen2016weakly} to mine large-scale category concepts in an open-vocabulary manner.
The fused proposal features $X^\mathrm{fuse}$ are forked into two fully-connected layers parallel, namely classification stream $W^c \in \mathcal{R}^{D \times C}$ and detection stream $W^d \in \mathcal{R}^{D \times C}$, producing two score matrices ${S}^{c}, {S}^{d} \in \mathcal{R}^{R \times C}$ respectively, where $C$ and $R$ denote the number of categories and proposals during training in image $I$,  respectively.
Different to work in~\cite{bilen2016weakly}, we adapt text embedding $T \in \mathcal{R}^{D \times C}$ of category names as the parameters $W^c$ of classification stream so that it imposes explicit visual-semantic constraint during MIL optimization.
At the same time, the detection stream is still learnable, since it focuses on localizing the foreground proposals, which is expected to be category-agnostic.
Thus, the two score matrices are computed as: $ S^c = \frac{ X^\mathrm{fuse} }{ \lVert  X^\mathrm{fuse} \rVert } \frac{ T }{ \lVert  T \rVert  }$ and $S^d = X^\mathrm{fuse} W^d$.
Then, both score matrices are normalized by softmax functions $\sigma(\cdot)$ over category and proposal axes, respectively.
The final score ${S}$ for assigning category $c$ to region $r$ is computed via an element-wise product: ${S} = \sigma({S}^{c}) \odot \sigma( ({S}^{d})^T )^T \in [0,1]$.
To acquire image-level classification scores for training, $S$ is summed for all regions $\varphi_{c} = \sum_{r=1}^{R} S_{rc}$.
Then the object-mining objective function is binary cross-entropy loss:
\begin{equation}
    \mathcal{L}_\mathrm{OM} = \sum_{c=1}^{C} {y}_{c}\log(\varphi_{c}) + (1 - {y}_{c})\log(1 - \varphi_{c})
    ,
    \label{eq:object_mining_loss}
\end{equation}
where ${y} \in \{0,1\}^{C}$ is the category one-hot label indicating image-level existence of category $c$.
Recently, WSOD works~\cite{tang2017multiple,tang2018pcl} also explicitly assign pseudo labels from the above mining module to learn more discriminative classifiers, which are also called instance refinement modules.
Thus, we also develop multiple open-vocabulary classification heads which uses a shared vision-language representation space to refine discovered object.
In addition, to reduce miss-localization, for each refinement branch we regress the bounding boxes which need high-quality proposals to provide PGT boxes as the previous WSOD methods~\cite{yang2019towards,ren2020instance}.
Therefore, the PGT boxes mined by the object mining module are used as box prompt input to SAM to refine boxes to supervise the first refinement branch, and the former refinement branch supervises the latter one.
Thus, the objective function of this multi-branch refinement is the sum of classification and regression losses over all branches: 
\begin{equation}
    \mathcal{L}_\mathrm{IR} = \sum_{k=1}^{K} \mathcal{L}_\mathrm{cls}^k + \mathcal{L}_\mathrm{reg}^k ,
    \label{eq:object_refinement_loss}
\end{equation}
We concatenate the text embedding $T$ with a background zero-vector as the classifier parameters $W^r \in \mathcal{R}^{D \times (C+1) }$ of refinement branch $k$.
The classification loss is defined as:
\begin{equation}
    \mathcal{L}_\mathrm{cls}^k = \sum_{r=1}^{R} \sum_{c=1}^{C+1} w^{k}_{c}  \hat{y}^k_{r,c} \log S_{r,c}^{k}
    ,
    \label{eq:refine_loss}
\end{equation}
where $w^{k}_{c}$ is the weight factor to smooth the learning process following~\cite{tang2017multiple}, $S^k \in \mathcal{R}^{R \times (C+1)}$ is computed by $ \frac{ X^\mathrm{fuse} }{ \lVert  X^\mathrm{fuse} \rVert } \frac{ W^r }{ \lVert  W^r \rVert  }$ and $\hat{y}^k_{r,c}$ is the PGT labels of proposal $r$ for category $c$ in branch $k$.
And $\mathcal{L}_\mathrm{reg}^k$ is the smooth L1 loss~\cite{NIPS2015_14bfa6bb} in branch $k$.

\section{Experiments}
\label{sec:exp}
\textbf{Datasets.}
We evaluate the proposed WSOVOD framework on Pascal VOC 2007, 2012~\cite{everingham2010Pascal} and MS COCO~\cite{lin2014microsoft}, which are widely used for WSOD.
In addition, we also use ILSVRC~\cite{ILSVRC} and LVIS~\cite{Gupta2019} for open-vocabulary learning, both of which are widely used for FSOVOD.
\begin{table*}[t]
    \caption{
        Comparison with the state-of-the-art OVOD methods on MS COCO and Pascal VOC.
    }
    \footnotesize
    \begin{center}
        \begin{adjustbox}{max width=0.8\textwidth}
            \begin{tabular}{l|c|c|c|ccc|c}
                \toprule

                \multirow{2}{*}{Method}&\multirow{2}{*}{Bakcbone}&\multicolumn{2}{|c|}{Train Supervision}&\multicolumn{3}{|c|}{COCO}&VOC07\\

                \cmidrule{3-8}
                
                &&Image-level&Object-level&{${AP_N}$}&{${AP_B}$}&{${AP}$}&$m$AP\\
                
                \midrule
                
                ZS-LAB\hfilll\cite{bansal2018zero}&Incept.-Res. v2&--&COCO 48 cls.&0.3&\textcolor{lightgray}{24.9}&\textcolor{lightgray}{--}&--\\

                DELO\hfilll\cite{zhu2020don}&DarkNet19&--&COCO 48 cls.&3.4&\textcolor{lightgray}{--}&\textcolor{lightgray}{13.0}&--\\

                PL\hfilll\cite{Rahman}&RN50-FPN&--&COCO 48 cls.&4.1&\textcolor{lightgray}{35.9}&\textcolor{lightgray}{7.4}&--\\

                SAN\hfilll\cite{rahman2020zero}&RN50&--&COCO 48 cls.&2.6&\textcolor{lightgray}{13.9}&\textcolor{lightgray}{4.3}&--\\

                BLC\hfilll\cite{zheng2020background}&RN50&--&COCO 48 cls.&4.5&\textcolor{lightgray}{42.1}&\textcolor{lightgray}{8.2}&--\\

                MS-Zero\hfilll\cite{gupta2020multi}&RN101&--&COCO 48 cls.&--&\textcolor{lightgray}{--}&\textcolor{lightgray}{$\mathbf{30.7}$}&--\\

                ContrastZSD\hfilll\cite{yan2022semantics}&RN101&--&COCO 48 cls.&6.3&\textcolor{lightgray}{45.1}&\textcolor{lightgray}{11.1}&--\\

                SSB\hfilll\cite{khandelwal2023frustratingly}&RN101&--&COCO 48 cls.&10.2&\textcolor{lightgray}{$\mathbf{48.9}$}&\textcolor{lightgray}{16.9}&--\\

                RRFS\hfilll\cite{huang2022robust}&RN101&--&COCO 48 cls.&$\mathbf{13.4}$&\textcolor{lightgray}{42.3}&\textcolor{lightgray}{20.4}&--\\

                \midrule

                OVR-CNN\hfilll\cite{zareian2021open}&RN50-C4&COCO caption&COCO 48 cls.&22.8&\textcolor{lightgray}{46.0}&\textcolor{lightgray}{39.9} & 52.9\\

                ViLD\hfilll\cite{gu2021open}&RN50-FPN&CLIP400M&COCO 48 cls.&27.6&\textcolor{lightgray}{59.5}&\textcolor{lightgray}{51.3}&--\\
                
                ZSD-YOLO\hfilll\cite{xie2022zero}&CSP-DarkNet53&CLIP400M&COCO 48 cls.&13.6&\textcolor{lightgray}{31.7}&\textcolor{lightgray}{19.0}&--\\

                HierKD\hfilll\cite{ma2022open}&RN50-FPN&Conceptual Captions&COCO 48 cls.&20.3&\textcolor{lightgray}{51.3}&\textcolor{lightgray}{43.2}&--\\

                Detic\hfilll\cite{zhou2022detecting}&RN50-C4&COCO caption&COCO 48 cls.&27.8&\textcolor{lightgray}{47.1}&\textcolor{lightgray}{45.0}&--\\

                Detic\hfilll\cite{zhou2022detecting}&RN50-C4&--&COCO 48 cls.&1.3&\textcolor{lightgray}{--}&\textcolor{lightgray}{39.3}&--\\

                RKDWTF\hfilll\cite{bangalath2022bridging}&RN50-C4&COCO caption&COCO 48 cls.&36.6&\textcolor{lightgray}{54.0}&\textcolor{lightgray}{49.4}&--\\

                SGDN\hfilll\cite{shi2023open}&RN50&Flickr30K, Visual Genome&COCO 48 cls.&$\mathbf{37.5}$&\textcolor{lightgray}{$\mathbf{61.0}$}&\textcolor{lightgray}{$\mathbf{54.9}$}&--\\

                PBBL\hfilll\cite{Gao2021b}&RN50&COCO Caption, Visual Genome, SBU Caption&COCO 48 cls.&30.8&\textcolor{lightgray}{46.1}&\textcolor{lightgray}{42.1}&59.2\\

                \midrule
                
                WSOVOD&RN50-WS-MRRP&VOC07&--&15.4&\textcolor{lightgray}{7.8}&\textcolor{lightgray}{9.8}&63.4\\

                WSOVOD&RN50-WS-MRRP&VOC12&--&17.0&\textcolor{lightgray}{9.3}&\textcolor{lightgray}{11.3}&\textbf{64.8}\\

                WSOVOD&RN50-WS-MRRP&ILSVRC&--&9.1&\textcolor{lightgray}{6.4}&\textcolor{lightgray}{7.0}&26.7\\

                WSOVOD&RN50-WS-MRRP&LVIS&--&16.7&\textcolor{lightgray}{11.0}&\textcolor{lightgray}{13.2}&31.0\\

                WSOVOD&RN50-WS-MRRP&COCO&--&$\mathbf{35.0}$&\textcolor{lightgray}{$\mathbf{27.9}$}&\textcolor{lightgray}{$\mathbf{29.8}$}&60.9\\

                \midrule

                WSOVOD&RN50-WS-MRRP&VOC07, VOC12&--&19.2&\textcolor{lightgray}{12.4}&\textcolor{lightgray}{15.1}&\textbf{65.0}\\

                WSOVOD&RN50-WS-MRRP&COCO, VOC07, VOC12&--&35.4&\textcolor{lightgray}{27.3}&\textcolor{lightgray}{29.8}&\textbf{65.0}\\

                WSOVOD&RN50-WS-MRRP&COCO, ILSVRC&--&35.6&\textcolor{lightgray}{27.7}&\textcolor{lightgray}{30.0}&61.2\\

                WSOVOD&RN50-WS-MRRP&COCO, LVIS&--&$\mathbf{36.7}$&\textcolor{lightgray}{$\mathbf{28.4}$}&\textcolor{lightgray}{$\mathbf{30.3}$}&62.3\\

                \bottomrule
            \end{tabular}
        \end{adjustbox}
    \end{center}
    \label{table_open_vocabulary_and_mixed_datasets}
\end{table*}

\begin{table}[t]
    \caption{
        LVIS experiments with RN18 on VOC07 and MS COCO. (``*'' refers to sampling by BCAS.)
    }
    \footnotesize
    \begin{center}
        \begin{adjustbox}{max width=0.45\textwidth}
            \begin{tabular}{c|c|cc|ccc}
                \toprule
                \multirow{4}{*}{Method}&\multirow{4}{*}{Train}&\multicolumn{5}{|c}{\multirow{1}{*}{Test}}\\
                \cmidrule{3-7}
                
                &&\multicolumn{2}{|c|}{PASCAL VOC 2007}&\multicolumn{3}{c}{MS COCO}\\
                
                &&\multirow{2}{*}{$m$AP}&\multirow{2}{*}{CorLoc}&\multicolumn{3}{c}{Avg. Precision, IoU:}\\
                
                &&&&0.5:0.95&0.5&0.75\\
                
                \midrule
                
                WSOVOD&LVIS&31.0&44.5&4.8&12.9&5.9\\

                WSOVOD*&LVIS&31.7&47.7&6.6&16.7&7.8\\

                WSOVOD&COCO&60.5&78.2&20.1&29.7&21.2\\
                
                WSOVOD&LVIS, COCO& $\mathbf{61.7}$ & $\mathbf{79.3}$ & $\mathbf{21.0}$ & $\mathbf{30.1}$ & $\mathbf{22.2}$ \\
                \bottomrule
            \end{tabular}
        \end{adjustbox}
    \end{center}
    \label{table_lvis}
\end{table}

\begin{table*}[t]
    \caption{
        Comparison with the state-of-the-art WSOD methods on PASCAL VOC 2007, 2012 and MS COCO.
    }
    \footnotesize
    \begin{center}
        \begin{adjustbox}{max width=0.8\textwidth}
            \begin{tabular}{l|c|c|cc|cc|ccc}
                \toprule

                \multirow{3}{*}{Method}&\multirow{3}{*}{Supervision}&\multirow{3}{*}{Bakcbone}&\multicolumn{2}{c|}{VOC 2007}&\multicolumn{2}{c|}{VOC 2012}&\multicolumn{3}{c}{MS COCO}\\

                &&&\multirow{2}{*}{$m$AP}&\multirow{2}{*}{CorLoc}&\multirow{2}{*}{$m$AP}&\multirow{2}{*}{CorLoc}&\multicolumn{3}{c}{Avg. Precision, IoU:}\\

                &&&&&&&0.5:0.95&0.5&0.75\\
                
                \midrule
                WSDDN\hfilll\cite{bilen2016weakly}&$\mathcal{I}$&VGG16&34.8&53.5&--&--&9.5&19.2&8.2\\
                ContextLocNet\hfilll\cite{kantorov2016contextlocnet}&$\mathcal{I}$&VGG-F&36.3&55.1&35.3&54.8&11.1&22.1&10.7\\
                OICR\hfilll\cite{tang2017multiple}&$\mathcal{I}$&VGG16&41.2&60.6&37.9&62.1&--&--&--\\
                
                ML-LocNet\hfilll\cite{zhang2018ml}&$\mathcal{I}$&VGG16&48.4&67.0&42.2&66.3&--&16.2&--\\
                
                MELM\hfilll\cite{wan2018min}&$\mathcal{I}$&VGG16&47.3&61.4&42.4&--&--&--&--\\

                C-MIL\hfilll\cite{wan2019c}&$\mathcal{I}$&VGG16&50.5&65.0&46.7&67.4&--&--&--\\
                PCL\hfilll\cite{tang2018pcl}&$\mathcal{I}$&VGG16&43.5&--&--&--&8.5&19.4&--\\
                WSOD$^2$\hfilll\cite{zeng2019wsod2}&$\mathcal{I}$&VGG16&53.6&69.5&47.2&71.9&10.8&22.7&--\\

                C-MIDN\hfilll\cite{gao2019c}&$\mathcal{I}$&VGG16&52.6&68.7&50.2&71.2&9.6&21.4&--\\
                MIST\hfilll\cite{ren2020instance}&$\mathcal{I}$&VGG16&54.9&68.8&52.1&70.9&12.4&25.8&10.5\\

                PG-PS\hfilll\cite{cheng2020high}&$\mathcal{I}$&VGG16&51.1&69.2&48.3&68.7&--&20.7&--\\
                UWSOD\hfilll\cite{shen2020uwsod}&$\mathcal{I}$&RN18-WS-MRRP&45.0&63.8&46.2&65.7&3.1&10.1&1.4\\

                SPE\hfilll\cite{liao2022end}&$\mathcal{I}$&CaiT&51.0&70.4&--&--&7.2&18.2&4.8\\

                Seo et al.\hfilll\cite{seo2022object}&$\mathcal{I}$&RN101&58.7&69.8&56.2&71.2&14.4&29.0&12.4\\                

                \midrule
                
                WSOVOD\hfill&$\mathcal{I}$&VGG16& 59.1&77.2&59.8&79.7&18.8&27.1&19.7\\

                WSOVOD\hfill&$\mathcal{I}$&RN18-WS-MRRP&63.0&$\mathbf{80.6}$&61.9&$\mathbf{81.0}$&20.1&$\mathbf{29.7}$&21.2\\

                WSOVOD\hfill&$\mathcal{I}$&RN50-WS-MRRP&$\mathbf{63.4}$&80.1&$\mathbf{62.1}$&80.7&$\mathbf{20.5}$&29.1&$\mathbf{21.4}$\\

               \midrule
                Fast RCNN\hfill\cite{girshick2015fast}&$\mathcal{O}$&VGG16&66.9&--&65.7&--&18.9&38.6&--\\
             
                Faster RCNN\hfill\cite{NIPS2015_14bfa6bb}&$\mathcal{O}$&VGG16&69.9&--&67.0&--&21.2&41.5&--\\

                \midrule
                
                WSOVOD\hfill&$\mathcal{B}$&VGG16&67.2&88.2&65.4&84.5&16.4&31.1&15.3\\

                WSOVOD\hfill&$\mathcal{B}$&RN18-WS-MRRP&68.8&90.9&66.3&89.2&19.8&37.6&18.5\\

                WSOVOD\hfill&$\mathcal{B}$&RN50-WS-MRRP& $\mathbf{71.8}$ & $\mathbf{91.0}$ & $\mathbf{69.7}$ & $\mathbf{90.0}$ & $\mathbf{21.6}$ & $\mathbf{40.6}$ & $\mathbf{20.8}$ \\
                \bottomrule
            \end{tabular}
        \end{adjustbox}
    \end{center}
    \label{table_voc2007voc2012mscoco_wsod_sota}
\end{table*}

\textbf{Evaluation Metrics.}
Following the common setting of FSOVOD, we also split COCO into $17$ novel classes and $48$ base classes, and use $AP_N$ and $AP_B$ to evaluate the results of $17$ novel classes and $48$ base classes, respectively.
We also use $AP$ to evaluate the results of $17+48$ total classes.
To compare the model performance in the WSOD setting, we use two evaluation metrics: CorLoc and $m$AP.
Correct localization~(CorLoc) is a commonly-used measurement that quantifies the localization performance by the percentage of images that contain at least one object instance with at least $50\%$ IoU to the ground-truth boxes.
Mean average precision~($m$AP) follows standard Pascal VOC protocol to report the $m$AP at $50\%$ IoU of the detected boxes with the ground-truth ones.
Furthermore, we report standard COCO metrics for WSOD, including AP at different IoU thresholds.
\textbf{Implementation Details.}
We use VGG16~\cite{simonyan2014very}, RN18/50-WS-MRRP~\cite{shen2020enabling}, initialized with the weights pre-trained on ILSVRC as vision backbones.
We use synchronized SGD training on Nvidia 3090 with a batch size of $4$, a mini-batch involving $1$ images per GPU.
We use learning rates of $1e^{-3}$ and $1e^{-2}$ for VGG16 and RN18/50-WS-MRRP backbone, respectfully, a momentum of $0.9$, a dropout rate of $0.5$, a learning rate decay weight of $0.1$.
We fix the backbone weights for WSOD but set a $1e^{-5}$ learning rate to backbones for OVOD.
\subsection{Open-Vocabulary and Cross-Dataset Detection}
\label{sec:exp_ovcdod}
Since we are the first exploration for WSOVOD, we compare the proposed WSOVOD framework with fully-supervised open-vocabulary object detection~(FSOVOD).
Noted that FSOVOD divides the MS COCO categories into $48$ base and $17$ novel classes~\cite{bansal2018zero}, and uses object-level annotations of $48$ base classes during training.
In addition, in order to expand vocabulary learning, some works~\cite{zareian2021open,zhou2022detecting,zareian2021open} use weak annotation information including novel classes, such as tags, captions, and etc.
The first and second parts of Tab.\,\ref{table_open_vocabulary_and_mixed_datasets} shows the state-of-the-art FSOVOD results without and with image-level annotation, respectively.
The $6th$ row in the second part removes COCO captions in Detic~\cite{zhou2022detecting}, which results in a dramatic performance drop in novel classes with only $1.3\%$ $AP_N$.
This shows that fully-supervised methods are hard to generalize well to detect novel classes if they lack the supervision information of a large vocabulary.
Therefore, it is necessary to study WSOVOD on large-vocabulary datasets with only category annotations.
As shown in the third part of Tab.~\ref{table_open_vocabulary_and_mixed_datasets}, WSOVOD exhibits strong generalization ability despite large differences between train and test distributions.
In particular, WSOVOD significantly improves the $AP_N$ performance of novel classes compared to FSOVOD with only object-level supervision.
On COCO novel classes, WSOVOD even surpasses FSOVOD methods, which require both image-level and object-level supervision.
We further conduct experiments to train our WSOVOD with multiple datasets jointly in the bottom part of Tab.~\ref{table_open_vocabulary_and_mixed_datasets}.
We observe that:
(1) Cross-dataset learning achieves superior or at least comparable results to the corresponding single dataset.
For instance, combing VOC07 and VOC12 significantly improves the COCO AP$_N$ with gains of $3.8\%$ and $2.2\%$ compared to separately using VOC07 and VOC12 datasets, respectively.
(2) Adding more image-level concepts to COCO further improves the COCO AP$_N$.
For instance, adding ILSVRC to COCO performs better than adding VOC07 and VOC12 to COCO.
(3) Adding denser image-level annotations significantly improves results.
For example, LVIS and COCO share the same training set, and directly combining LVIS and COCO improves $1.7\%$ AP$_N$, although the image-level labels in LVIS are incomplete.

\subsection{Rescuing Federated and Long-Tail Data}
\label{sec:exp_lvis}
To better adjust and analyze our WSOVOD, we further conduct experiments on the difficult federated and long-tail distribution LVIS~\cite{Gupta2019} dataset, as shown in Tab.~\ref{table_lvis}.
When only using LVIS for training, the performance of WSOVOD reaches saturation around $1$ epoch on VOC07 and COCO.
This is because LVIS is a federated dataset with sparse annotations.
So, image-level labels are not exhaustively annotated with all classes.
The missing classes are treated as background and generate incorrect supervision signals.
To this end, we introduce a batch-class-aware sampling, termed BCAS.
In BCAS, the data sampler first picks a category and then selects multiple images containing that category to form a mini-batch.
When equipped with BCAS for LVIS, WSOVOD reaches saturation at about $3$ epochs and improves $3.8\%$ AP$_{0.5}$ on COCO.
We further add COCO to LVIS training without BCAS and observe substantial performance improvements on VOC 2007 with gains of $30.7\%$ $mAP$ and $34.8\%$ CorLoc, respectively.
Compared to single COCO, combing LVIS with COCO also significantly improves the VOC 2007 $mAP$ from $60.5\%$ to $61.7\%$ and CorLoc from $78.2\%$ to $79.3\%$, respectively. 
This reveals that incomplete image-level annotated data is helpful for WSOVOD cross-dataset training.
\subsection{Weakly Supervised Object Detection}
\label{sec:exp_stoa_wsod}
We compare our proposed method with the state-of-the-art WSOD methods on VGG16, RN18-WS-MRRP, and RN50-WS-MRRP backbones.
Tab.\,\ref{table_voc2007voc2012mscoco_wsod_sota} shows the performance comparisons on the Pascal VOC 2007, Pascal VOC 2012, and MS COCO, where $\mathcal{I}$, $\mathcal{O}$ and $\mathcal{B}$ denote image-level supervision, object-level supervision with class labels, and object-level supervision without class labels, respectively.
With the VGG16 backbone, the proposed WSOVOD suppresses the performances of all previous WSOD methods for $mAP$ and CorLoc on Pascal VOC and $AP_{0.5:0.95}$ on MS COCO.
The proposed WSOVOD with RN18-WS-MRRP backbone reaches the new state-of-the-art of $80.6\%$ and $81.0\%$ CorLoc on Pascal VOC 2007 and 2012, respectively, and $29.7\%$ AP$_{0.5}$ on MS COCO.
With RN50-WS-MRRP backbone, WSOVOD further sets new state-of-the-art of $63.4\%$ and $62.1\%$ $mAP$ on Pascal VOC 2007 and VOC 2012, respectively, and $20.5\%$ $AP_{0.5:0.95}$ and $21.4\%$ AP$_{0.75}$ on MS COCO.
Furthermore, with object-level supervision without class labels, our proposed WSOVOD even shows comparable performance compared to FSOD in all datasets.
\subsection{Ablation Study}
\label{sec:exp_ab}
\begin{table}[t]
    \caption{
        Ablation study of DAFE with RN18-WS-MRRP backbone on VOC 2007.
    }
    \footnotesize
    \begin{center}
        \begin{adjustbox}{max width=0.4\textwidth}
            \begin{tabular}{l|cc|cc}
                \toprule
                
                \multirow{2}{*}{Train Data}&\multicolumn{2}{c|}{without DAFE}&\multicolumn{2}{c}{with DAFE}\\

                &$m$AP&CorLoc&$m$AP&CorLoc\\

                \midrule
                
                VOC07 & 62.6 & 78.7 & 63.0 {\color{green} ($\uparrow$ 0.4) } & 80.6 {\color{green} ($\uparrow$ 1.9) }\\
                
                VOC07, VOC12 & 63.5 & 79.2 & 64.1 {\color{green} ($\uparrow$ 0.6) } & 82.2 {\color{green} ($\uparrow$ 3.0) }\\
                VOC07, COCO & 61.4 & 78.2 & 63.0 {\color{green} ($\uparrow$ 1.6) } & 80.5 {\color{green} ($\uparrow$ 2.3) }\\

                \bottomrule
            \end{tabular}
        \end{adjustbox}
    \end{center}
    \label{table_ablation_data_aware}
\end{table}

\begin{table}[t]
    \caption{
        Ablation study of proposal generator with RN18-WS-MRRP backbone on VOC 2007.
    }
    \footnotesize
    \begin{center}
        \begin{adjustbox}{max width=0.4\textwidth}
            \begin{tabular}{l|cc}
                \toprule

                Proposal &m$AP$&CorLoc\\

                \midrule

                LO-WSRPN& 46.7 {\color{white} ($\uparrow$ 0.00) } & 65.1 {\color{white} ($\uparrow$ 0.00) } \\
                
                MCG& 54.2 {\color{green} ($\uparrow$ 7.50) } & 71.9 {\color{green} ($\uparrow$ 6.80) } \\

                SAM& 61.7 {\color{green} ($\uparrow$ 15.0) } & 77.5 {\color{green} ($\uparrow$ 12.4) } \\

                LO-WSRPN + SAM& 62.5 {\color{green} ($\uparrow$ 15.8) } & 79.9 {\color{green} ($\uparrow$ 14.8) } \\

                LO-WSRPN + SAM + refine& 63.0 {\color{green} ($\uparrow$ 16.3) } & 81.0 {\color{green} ($\uparrow$ 15.9) } \\

                \bottomrule
            \end{tabular}
        \end{adjustbox}
    \end{center}
    \label{table_ablation_rpn}

\end{table}

\begin{table}[t]
    \caption{
        Ablation study of proposal generator recall on VOC 2007 \textit{trainval} on IoU$_{0.5}$ / average IoU$_{0.5:1.0}$.
    }
    \begin{center}
        \begin{adjustbox}{max width=0.49\textwidth}
            \begin{tabular}{l|ccccc}
                \toprule
                Method&AR@1&AR@10&AR@100&AR@1,000&AR@2,000\\
                 
                \midrule
                LO-WSRPN&8.1/2.3&25.1/8.1&53.2/20.7&74.7/32.3&74.9/32.4\\
                MCG&$\mathbf{11.9}$/5.5&$\mathbf{39.1}$/18.0&70.0/37.6&89.5/54.8&93.0/58.8\\

                SAM&8.7/$\mathbf{6.1}$&35.3/$\mathbf{24.9}$&$\mathbf{80.9}$/$\mathbf{52.9}$&$\mathbf{92.1}$/$\mathbf{59.1}$&92.1/59.1\\

                LO-WSRPN+SAM&8.1/2.3&25.1/8.1&53.8/21.0&86.2/46.5&97.1/63.2\\

                LO-WSRPN+SAM+refine&8.5/4.6&30.7/11.2&65.5/32.7&88.4/52.5&$\mathbf{98.0}$/$\mathbf{65.2}$\\
                
                \bottomrule
            \end{tabular}
        \end{adjustbox}
    \end{center}
    \label{table_ablation_recall}
\end{table}

We conducted three sets of ablation studies to investigate the effectiveness of the proposed modules.
We firstly ablate data-aware features extractor~(DAFE) in Tab.\,\ref{table_ablation_data_aware} to verify the effectiveness of DAFE for training multiple datasets.
We test all models on VOC07 \textit{test}.
When training on VOC12 and VOC07, DAFE improves $m$AP by $0.6\%$ and CorLoc by $3.0\%$.
Thus, DAFE significantly improves the detection and localization performance, indicating that DAFE is simple and effective.
When training on COCO and VOC07, DAFE improves $m$AP by $1.6\%$ and CorLoc by $2.3\%$.
It demonstrates that DAFE also deals well with the large distribution gap.
DAFE also performs well on a single dataset. 
Thus, introducing global image-level context to local proposal-level features is helpful to WSOD. 
This reveals that DAFE not only successfully gathers dataset-level bias but also image-level context.
Secondly, we ablate the proposal generator in Tab.\,\ref{table_ablation_rpn}.
It shows that, as observed in~\cite{shen2020uwsod}, only using predictions from the model itself as supervision results in noisy training, so that $m$AP and CorLoc only reach $46.7\%$ and $65.1\%$, respectively.
When using traditional object proposals, such as MCG~\cite{pont2016multiscale}, the performance is significantly improved, but compared with SAM proposals based on high-level semantic information, it is still much worse.
When adding SAM proposals to LO-WSRPN proposals with the refinement mechanism, our method improves $m$AP and CorLoc by $16.3\%$ and $15.9\%$, respectively.
Thirdly, we ablate the proposal generator on VOC07 \textit{trainval} in terms of proposal recall on IoU$_{0.5}$ and average IoU$_{0.5:1.0}$ in Tab.\,\ref{table_ablation_recall}.
We find that SAM provides high-quality proposals, but it has too few proposals that are less than $1,000$.
The recall of proposals provided by traditional object proposals, \textit{i.e.}, MCG, is relatively low on AR@$100$.
And our LO-WSRPN achieves the highest AR@$2,000$ with $98.0\%$ under IoU$_{0.5}$ and $65.2$ under IoU$_{0.5:1.0}$.

\section{Conclusion}
\label{sec:conclusion}
In this paper, we propose a weakly supervised open-vocabulary object detection framework, namely WSOVOD, which extends WSOD to detect and localize open-vocabulary concepts and utilizes diverse and large-scale datasets with only image-level annotation.
On the widely used benchmark Pascal VOC and MS COCO, the proposed WSOVOD achieves on-par or even better performance compared to FSOVOD methods with box-level supervision.
Moreover, WSOVOD sets new state-of-the-art results for WSOD in both localization and detection tasks.
\section{Acknowledgments}
This work was supported by National Key R$\&$D Program of China~(No.2022ZD0118202), the National Science Fund for Distinguished Young Scholars~(No.62025603), the National Natural Science Foundation of China~(No.U21B2037, No.U22B2051, No.62176222, No.62176223, No. 2176226, No.62072386, No.62072387, No.62072389, No.62002305, NO. 62102151 and No.62272401), the Natural Science Foundation of Fujian Province of China~(No.2021J01002, No.2022J06001), and CCF-Tencent Rhino-Bird Open Research Fund.
\bibliography{wsovod}

\begin{thebibliography}{69}
\providecommand{\natexlab}[1]{#1}

\bibitem[{Bahng et~al.(2022)Bahng, Jahanian, Sankaranarayanan, and Isola}]{Bahng2022}
Bahng, H.; Jahanian, A.; Sankaranarayanan, S.; and Isola, P. 2022.
\newblock {Exploring Visual Prompts for Adapting Large-Scale Models}.
\newblock \emph{arXiv}.

\bibitem[{Bangalath et~al.(2022)Bangalath, Maaz, Khattak, Khan, and Shahbaz~Khan}]{bangalath2022bridging}
Bangalath, H.; Maaz, M.; Khattak, M.~U.; Khan, S.~H.; and Shahbaz~Khan, F. 2022.
\newblock Bridging the gap between object and image-level representations for open-vocabulary detection.
\newblock In \emph{NeurIPS}.

\bibitem[{Bansal et~al.(2018)Bansal, Sikka, Sharma, Chellappa, and Divakaran}]{bansal2018zero}
Bansal, A.; Sikka, K.; Sharma, G.; Chellappa, R.; and Divakaran, A. 2018.
\newblock Zero-Shot Object Detection.
\newblock In \emph{ECCV}.

\bibitem[{Bilen and Vedaldi(2016)}]{bilen2016weakly}
Bilen, H.; and Vedaldi, A. 2016.
\newblock Weakly Supervised Deep Detection Networks.
\newblock In \emph{CVPR}.

\bibitem[{Carion et~al.(2020)Carion, Massa, Synnaeve, Usunier, Kirillov, and Zagoruyko}]{carion2020end}
Carion, N.; Massa, F.; Synnaeve, G.; Usunier, N.; Kirillov, A.; and Zagoruyko, S. 2020.
\newblock End-To-End Object Detection with Transformers.
\newblock In \emph{ECCV}.

\bibitem[{Cheng et~al.(2020)Cheng, Yang, Gao, Guo, and Han}]{cheng2020high}
Cheng, G.; Yang, J.; Gao, D.; Guo, L.; and Han, J. 2020.
\newblock High-Quality Proposals for Weakly Supervised Object Detection.
\newblock \emph{TIP}.

\bibitem[{Dietterich, Lathrop, and Lozano-P{\'e}rez(1997)}]{dietterich1997solving}
Dietterich, T.~G.; Lathrop, R.~H.; and Lozano-P{\'e}rez, T. 1997.
\newblock Solving the Multiple Instance Problem with Axis-Parallel Rectangles.
\newblock \emph{AI}.

\bibitem[{Doll{\'{a}}r and Zitnick(2013)}]{Dollar2013}
Doll{\'{a}}r, P.; and Zitnick, C.~L. 2013.
\newblock {Structured Forests for Fast Edge Detection}.
\newblock In \emph{ICCV}.

\bibitem[{Everingham et~al.(2010)Everingham, Van~Gool, Williams, Winn, and Zisserman}]{everingham2010Pascal}
Everingham, M.; Van~Gool, L.; Williams, C.~K.; Winn, J.; and Zisserman, A. 2010.
\newblock The Pascal Visual Object Classes (voc) Challenge.
\newblock \emph{IJCV}.

\bibitem[{Felzenszwalb and Huttenlocher(2004)}]{Felzenszwalb2004}
Felzenszwalb, P.~F.; and Huttenlocher, D.~P. 2004.
\newblock {Efficient Graph-Based Image Segmentation}.
\newblock \emph{IJCV}.

\bibitem[{Feng et~al.(2022)Feng, Zhong, Jie, Chu, Ren, Wei, Xie, and Ma}]{Feng2022}
Feng, C.; Zhong, Y.; Jie, Z.; Chu, X.; Ren, H.; Wei, X.; Xie, W.; and Ma, L. 2022.
\newblock {PromptDet: Towards Open-Vocabulary Detection Using Uncurated Images}.
\newblock In \emph{ECCV}.

\bibitem[{Gao et~al.(2022)Gao, Xing, Niebles, Li, Xu, Liu, and Xiong}]{Gao2021b}
Gao, M.; Xing, C.; Niebles, J.~C.; Li, J.; Xu, R.; Liu, W.; and Xiong, C. 2022.
\newblock {Open Vocabulary Object Detection with Pseudo Bounding-Box Labels}.
\newblock In \emph{ECCV}.

\bibitem[{Gao et~al.(2019)Gao, Liu, Guo, Ye, Wan, You, and Fan}]{gao2019c}
Gao, Y.; Liu, B.; Guo, N.; Ye, X.; Wan, F.; You, H.; and Fan, D. 2019.
\newblock C-Midn: Coupled Multiple Instance Detection Network with Segmentation Guidance for Weakly Supervised Object Detection.
\newblock In \emph{ICCV}.

\bibitem[{Girshick(2015)}]{girshick2015fast}
Girshick, R. 2015.
\newblock Fast R-Cnn.
\newblock In \emph{CVPR}.

\bibitem[{Gu et~al.(2022)Gu, Lin, Kuo, and Cui}]{gu2021open}
Gu, X.; Lin, T.-Y.; Kuo, W.; and Cui, Y. 2022.
\newblock Open-Vocabulary Detection via Vision and Language Knowledge Distillation.
\newblock \emph{ICLR}.

\bibitem[{Gupta, Doll{\'{a}}r, and Girshick(2019)}]{Gupta2019}
Gupta, A.; Doll{\'{a}}r, P.; and Girshick, R. 2019.
\newblock {LVIS: A Dataset for Large Vocabulary Instance Segmentation}.
\newblock In \emph{CVPR}.

\bibitem[{Gupta et~al.(2020)Gupta, Anantharaman, Mamgain, Balasubramanian, Jawahar et~al.}]{gupta2020multi}
Gupta, D.; Anantharaman, A.; Mamgain, N.; Balasubramanian, V.~N.; Jawahar, C.; et~al. 2020.
\newblock A multi-space approach to zero-shot object detection.
\newblock In \emph{WCACV}.

\bibitem[{Huang et~al.(2022)Huang, Han, Cheng, and Zhang}]{huang2022robust}
Huang, P.; Han, J.; Cheng, D.; and Zhang, D. 2022.
\newblock Robust region feature synthesizer for zero-shot object detection.
\newblock In \emph{CVPR}.

\bibitem[{Jia et~al.(2009)Jia, Wei, Socher, Li, Li, and Li}]{deng2009large}
Jia, D.; Wei, D.; Socher, R.; Li, L.-J.; Li, K.; and Li, F.-F. 2009.
\newblock A Large-Scale Hierarchical Image Database.
\newblock In \emph{CVPR}.

\bibitem[{Jia et~al.(2022)Jia, Tang, Chen, Cardie, Belongie, Hariharan, and Lim}]{Jia2022}
Jia, M.; Tang, L.; Chen, B.-C.; Cardie, C.; Belongie, S.; Hariharan, B.; and Lim, S.-N. 2022.
\newblock {Visual Prompt Tuning}.
\newblock In \emph{ECCV}.

\bibitem[{Jiang et~al.(2022)Jiang, Zhu, Liu, Song, Li, and Min}]{Jiang2022}
Jiang, S.; Zhu, Y.; Liu, C.; Song, X.; Li, X.; and Min, W. 2022.
\newblock {Dataset Bias in Few-shot Image Recognition}.
\newblock \emph{TPAMI}.

\bibitem[{Kamath et~al.(2021)Kamath, Singh, Lecun, Synnaeve, Misra, and Carion}]{Kamath}
Kamath, A.; Singh, M.; Lecun, Y.; Synnaeve, G.; Misra, I.; and Carion, N. 2021.
\newblock {MDETR - Modulated Detection for End-To-End Multi-Modal Understanding}.
\newblock In \emph{ICCV}.

\bibitem[{Kantorov et~al.(2016)Kantorov, Oquab, Cho, and Laptev}]{kantorov2016contextlocnet}
Kantorov, V.; Oquab, M.; Cho, M.; and Laptev, I. 2016.
\newblock Contextlocnet: Context-Aware Deep Network Models for Weakly Supervised Localization.
\newblock In \emph{ECCV}.

\bibitem[{Khandelwal et~al.(2023)Khandelwal, Nambirajan, Siddiquie, Eledath, and Sigal}]{khandelwal2023frustratingly}
Khandelwal, S.; Nambirajan, A.; Siddiquie, B.; Eledath, J.; and Sigal, L. 2023.
\newblock Frustratingly Simple but Effective Zero-shot Detection and Segmentation: Analysis and a Strong Baseline.
\newblock \emph{arXiv}.

\bibitem[{Kim et~al.(2020)Kim, Lee, Jeong, and Kwak}]{kim2020tell}
Kim, D.; Lee, G.; Jeong, J.; and Kwak, N. 2020.
\newblock Tell Me What They're Holding: Weakly-Supervised Object Detection with Transferable Knowledge from Human-Object Interaction.
\newblock In \emph{AAAI}.

\bibitem[{Kim, Lee, and Choo(2021)}]{Kim2021}
Kim, E.; Lee, J.; and Choo, J. 2021.
\newblock {BiaSwap: Removing dataset bias with bias-tailored swapping augmentation}.
\newblock In \emph{ICCV}.

\bibitem[{Kirillov et~al.(2023)Kirillov, Mintun, Ravi, Mao, Rolland, Gustafson, Xiao, Whitehead, Berg, Lo, Doll{\'a}r, and Girshick}]{kirillov2023segany}
Kirillov, A.; Mintun, E.; Ravi, N.; Mao, H.; Rolland, C.; Gustafson, L.; Xiao, T.; Whitehead, S.; Berg, A.~C.; Lo, W.-Y.; Doll{\'a}r, P.; and Girshick, R. 2023.
\newblock Segment Anything.
\newblock In \emph{ICCV}.

\bibitem[{Li et~al.(2022)Li, Zhang, Zhang, Yang, Li, Zhong, Wang, Yuan, Zhang, Hwang et~al.}]{li2022grounded}
Li, L.~H.; Zhang, P.; Zhang, H.; Yang, J.; Li, C.; Zhong, Y.; Wang, L.; Yuan, L.; Zhang, L.; Hwang, J.-N.; et~al. 2022.
\newblock Grounded Language-Image Pre-Training.
\newblock In \emph{CVPR}.

\bibitem[{Liao et~al.(2022)Liao, Wan, Yao, Han, Zou, Wang, Feng, Yuan, and Ye}]{liao2022end}
Liao, M.; Wan, F.; Yao, Y.; Han, Z.; Zou, J.; Wang, Y.; Feng, B.; Yuan, P.; and Ye, Q. 2022.
\newblock End-to-End Weakly Supervised Object Detection with Sparse Proposal Evolution.
\newblock In \emph{ECCV}.

\bibitem[{Lin et~al.(2014)Lin, Maire, Belongie, Hays, Perona, Ramanan, Doll{\'a}r, and Zitnick}]{lin2014microsoft}
Lin, T.-Y.; Maire, M.; Belongie, S.; Hays, J.; Perona, P.; Ramanan, D.; Doll{\'a}r, P.; and Zitnick, C.~L. 2014.
\newblock Microsoft Coco: Common Objects in Context.
\newblock In \emph{ECCV}.

\bibitem[{Ma et~al.(2022{\natexlab{a}})Ma, Luo, Gao, Li, Chen, Wang, Zhang, and Hu}]{Ma2022}
Ma, Z.; Luo, G.; Gao, J.; Li, L.; Chen, Y.; Wang, S.; Zhang, C.; and Hu, W. 2022{\natexlab{a}}.
\newblock {Open-Vocabulary One-Stage Detection with Hierarchical Visual-Language Knowledge Distillation}.
\newblock In \emph{CVPR}.

\bibitem[{Ma et~al.(2022{\natexlab{b}})Ma, Luo, Gao, Li, Chen, Wang, Zhang, and Hu}]{ma2022open}
Ma, Z.; Luo, G.; Gao, J.; Li, L.; Chen, Y.; Wang, S.; Zhang, C.; and Hu, W. 2022{\natexlab{b}}.
\newblock Open-vocabulary one-stage detection with hierarchical visual-language knowledge distillation.
\newblock In \emph{CVPR}.

\bibitem[{Minderer et~al.(2022)Minderer, Gritsenko, Stone, Neumann, Weissenborn, Dosovitskiy, Mahendran, Arnab, Dehghani, Shen et~al.}]{minderer2022simple}
Minderer, M.; Gritsenko, A.; Stone, A.; Neumann, M.; Weissenborn, D.; Dosovitskiy, A.; Mahendran, A.; Arnab, A.; Dehghani, M.; Shen, Z.; et~al. 2022.
\newblock Simple open-vocabulary object detection.
\newblock In \emph{ECCV}.

\bibitem[{Pont-Tuset et~al.(2016)Pont-Tuset, Arbelaez, Barron, Marques, and Malik}]{pont2016multiscale}
Pont-Tuset, J.; Arbelaez, P.; Barron, J.~T.; Marques, F.; and Malik, J. 2016.
\newblock Multiscale Combinatorial Grouping for Image Segmentation and Object Proposal Generation.
\newblock \emph{TPAMI}.

\bibitem[{Radford et~al.(2021)Radford, Kim, Hallacy, Ramesh, Goh, Agarwal, Sastry, Askell, Mishkin, Clark et~al.}]{radford2021learning}
Radford, A.; Kim, J.~W.; Hallacy, C.; Ramesh, A.; Goh, G.; Agarwal, S.; Sastry, G.; Askell, A.; Mishkin, P.; Clark, J.; et~al. 2021.
\newblock Learning Transferable Visual Models from Natural Language Supervision.
\newblock In \emph{ICML}.

\bibitem[{Rahman, Khan, and Barnes(2020)}]{Rahman}
Rahman, S.; Khan, S.; and Barnes, N. 2020.
\newblock {Improved Visual-Semantic Alignment for Zero-Shot Object Detection}.
\newblock In \emph{AAAI}.

\bibitem[{Rahman, Khan, and Porikli(2020)}]{rahman2020zero}
Rahman, S.; Khan, S.~H.; and Porikli, F. 2020.
\newblock Zero-shot object detection: Joint recognition and localization of novel concepts.
\newblock \emph{IJCV}.

\bibitem[{Redmon et~al.(2016)Redmon, Divvala, Girshick, and Farhadi}]{Redmon_2016_CVPR}
Redmon, J.; Divvala, S.; Girshick, R.; and Farhadi, A. 2016.
\newblock You Only Look Once: Unified, Real-Time Object Detection.
\newblock In \emph{CVPR}.

\bibitem[{Redmon and Farhadi(2017)}]{8100173}
Redmon, J.; and Farhadi, A. 2017.
\newblock YOLO9000: Better, Faster, Stronger.
\newblock In \emph{CVPR}.

\bibitem[{Ren et~al.(2015)Ren, He, Girshick, and Sun}]{NIPS2015_14bfa6bb}
Ren, S.; He, K.; Girshick, R.; and Sun, J. 2015.
\newblock Faster R-CNN: Towards Real-Time Object Detection with Region Proposal Networks.
\newblock In \emph{NeurIPS}.

\bibitem[{Ren et~al.(2020)Ren, Yu, Yang, Liu, Lee, Schwing, and Kautz}]{ren2020instance}
Ren, Z.; Yu, Z.; Yang, X.; Liu, M.-Y.; Lee, Y.~J.; Schwing, A.~G.; and Kautz, J. 2020.
\newblock Instance-Aware, Context-Focused, and Memory-Efficient Weakly Supervised Object Detection.
\newblock In \emph{CVPR}.

\bibitem[{Russakovsky et~al.(2015)Russakovsky, Deng, Su, Krause, Satheesh, Ma, Huang, Karpathy, Khosla, Bernstein, Berg, and Fei-Fei}]{ILSVRC}
Russakovsky, O.; Deng, J.; Su, H.; Krause, J.; Satheesh, S.; Ma, S.; Huang, Z.; Karpathy, A.; Khosla, A.; Bernstein, M.; Berg, A.~C.; and Fei-Fei, L. 2015.
\newblock {ImageNet Large Scale Visual Recognition Challenge}.
\newblock \emph{IJCV}.

\bibitem[{Seo et~al.(2022)Seo, Bae, Sutherland, Noh, and Kim}]{seo2022object}
Seo, J.; Bae, W.; Sutherland, D.~J.; Noh, J.; and Kim, D. 2022.
\newblock Object discovery via contrastive learning for weakly supervised object detection.
\newblock In \emph{ECCV}.

\bibitem[{Sharma et~al.(2018)Sharma, Ding, Goodman, and Soricut}]{sharma2018conceptual}
Sharma, P.; Ding, N.; Goodman, S.; and Soricut, R. 2018.
\newblock Conceptual Captions: A Cleaned, Hypernymed, Image Alt-Text Dataset for Automatic Image Captioning.
\newblock In \emph{ACL}.

\bibitem[{Shen et~al.(2020{\natexlab{a}})Shen, Ji, Chen, Wu, and Huang}]{shen2020uwsod}
Shen, Y.; Ji, R.; Chen, Z.; Wu, Y.; and Huang, F. 2020{\natexlab{a}}.
\newblock UWSOD: Toward Fully-Supervised-Level Capacity Weakly Supervised Object Detection.
\newblock \emph{NeurIPS}.

\bibitem[{Shen et~al.(2020{\natexlab{b}})Shen, Ji, Wang, Chen, Zheng, Huang, and Wu}]{shen2020enabling}
Shen, Y.; Ji, R.; Wang, Y.; Chen, Z.; Zheng, F.; Huang, F.; and Wu, Y. 2020{\natexlab{b}}.
\newblock Enabling Deep Residual Networks for Weakly Supervised Object Detection.
\newblock In \emph{ECCV}.

\bibitem[{Shi, Hayat, and Cai(2023)}]{shi2023open}
Shi, H.; Hayat, M.; and Cai, J. 2023.
\newblock Open-Vocabulary Object Detection via Scene Graph Discovery.
\newblock In \emph{ACMMM}.

\bibitem[{Simonyan and Zisserman(2015)}]{simonyan2014very}
Simonyan, K.; and Zisserman, A. 2015.
\newblock Very deep convolutional networks for large-scale image recognition.
\newblock In \emph{ICLR}.

\bibitem[{Sun et~al.(2021)Sun, Cao, Yang, and Kitani}]{sun2021rethinking}
Sun, Z.; Cao, S.; Yang, Y.; and Kitani, K.~M. 2021.
\newblock Rethinking Transformer-Based Set Prediction for Object Detection.
\newblock In \emph{ICCV}.

\bibitem[{Tang et~al.(2018)Tang, Wang, Bai, Shen, Bai, Liu, and Yuille}]{tang2018pcl}
Tang, P.; Wang, X.; Bai, S.; Shen, W.; Bai, X.; Liu, W.; and Yuille, A. 2018.
\newblock Pcl: Proposal Cluster Learning for Weakly Supervised Object Detection.
\newblock \emph{TPAMI}.

\bibitem[{Tang et~al.(2017)Tang, Wang, Bai, and Liu}]{tang2017multiple}
Tang, P.; Wang, X.; Bai, X.; and Liu, W. 2017.
\newblock Multiple Instance Detection Network with Online Instance Classifier Refinement.
\newblock In \emph{CVPR}.

\bibitem[{Tian et~al.(2019)Tian, Shen, Chen, and He}]{tian2019fcos}
Tian, Z.; Shen, C.; Chen, H.; and He, T. 2019.
\newblock Fcos: Fully convolutional one-stage object detection.
\newblock In \emph{ICCV}.

\bibitem[{Torralba and Efros(2011)}]{Torralba2011}
Torralba, A.; and Efros, A.~A. 2011.
\newblock {Unbiased Look at Dataset Bias}.
\newblock In \emph{CVPR}.

\bibitem[{Uijlings et~al.(2013)Uijlings, Van De~Sande, Gevers, and Smeulders}]{uijlings2013selective}
Uijlings, J.~R.; Van De~Sande, K.~E.; Gevers, T.; and Smeulders, A.~W. 2013.
\newblock Selective Search for Object Recognition.
\newblock \emph{IJCV}.

\bibitem[{Wan et~al.(2019)Wan, Liu, Ke, Ji, Jiao, and Ye}]{wan2019c}
Wan, F.; Liu, C.; Ke, W.; Ji, X.; Jiao, J.; and Ye, Q. 2019.
\newblock C-Mil: Continuation Multiple Instance Learning for Weakly Supervised Object Detection.
\newblock In \emph{CVPR}.

\bibitem[{Wan et~al.(2018)Wan, Wei, Jiao, Han, and Ye}]{wan2018min}
Wan, F.; Wei, P.; Jiao, J.; Han, Z.; and Ye, Q. 2018.
\newblock Min-Entropy Latent Model for Weakly Supervised Object Detection.
\newblock In \emph{CVPR}.

\bibitem[{Xie and Zheng(2022)}]{xie2022zero}
Xie, J.; and Zheng, S. 2022.
\newblock Zero-shot Object Detection Through Vision-Language Embedding Alignment.
\newblock In \emph{ICDMW}.

\bibitem[{Yan et~al.(2022)Yan, Chang, Luo, Liu, Zhang, and Zheng}]{yan2022semantics}
Yan, C.; Chang, X.; Luo, M.; Liu, H.; Zhang, X.; and Zheng, Q. 2022.
\newblock Semantics-guided contrastive network for zero-shot object detection.
\newblock \emph{TPAMI}.

\bibitem[{Yang, Li, and Dou(2019)}]{yang2019towards}
Yang, K.; Li, D.; and Dou, Y. 2019.
\newblock Towards Precise End-To-End Weakly Supervised Object Detection Network.
\newblock In \emph{ICCV}.

\bibitem[{Zang et~al.(2022)Zang, Li, Zhou, Huang, and Loy}]{Zang2022}
Zang, Y.; Li, W.; Zhou, K.; Huang, C.; and Loy, C.~C. 2022.
\newblock {Open-Vocabulary DETR with Conditional Matching}.
\newblock In \emph{ECCV}.

\bibitem[{Zareian et~al.(2021)Zareian, Rosa, Hu, and Chang}]{zareian2021open}
Zareian, A.; Rosa, K.~D.; Hu, D.~H.; and Chang, S.-F. 2021.
\newblock Open-Vocabulary Object Detection Using Captions.
\newblock In \emph{CVPR}.

\bibitem[{Zeng et~al.(2019)Zeng, Liu, Fu, Chao, and Zhang}]{zeng2019wsod2}
Zeng, Z.; Liu, B.; Fu, J.; Chao, H.; and Zhang, L. 2019.
\newblock Wsod2: Learning Bottom-up and Top-down Objectness Distillation for Weakly-Supervised Object Detection.
\newblock In \emph{ICCV}.

\bibitem[{Zhang, Yang, and Feng(2018)}]{zhang2018ml}
Zhang, X.; Yang, Y.; and Feng, J. 2018.
\newblock Ml-Locnet: Improving Object Localization with Multi-View Learning Network.
\newblock In \emph{ECCV}.

\bibitem[{Zheng et~al.(2021)Zheng, Gao, Zhang, Li, Wang, Li, and Dong}]{zheng2020end}
Zheng, M.; Gao, P.; Zhang, R.; Li, K.; Wang, X.; Li, H.; and Dong, H. 2021.
\newblock End-To-End Object Detection with Adaptive Clustering Transformer.
\newblock In \emph{BMVC}.

\bibitem[{Zheng et~al.(2020)Zheng, Huang, Han, Huang, and Cui}]{zheng2020background}
Zheng, Y.; Huang, R.; Han, C.; Huang, X.; and Cui, L. 2020.
\newblock Background learnable cascade for zero-shot object detection.
\newblock In \emph{ACCV}.

\bibitem[{Zhou et~al.(2022{\natexlab{a}})Zhou, Yang, Loy, and Liu}]{Zhou2022a}
Zhou, K.; Yang, J.; Loy, C.~C.; and Liu, Z. 2022{\natexlab{a}}.
\newblock {Conditional Prompt Learning for Vision-Language Models}.
\newblock In \emph{CVPR}.

\bibitem[{Zhou et~al.(2022{\natexlab{b}})Zhou, Girdhar, Joulin, Kr{\"a}henb{\"u}hl, and Misra}]{zhou2022detecting}
Zhou, X.; Girdhar, R.; Joulin, A.; Kr{\"a}henb{\"u}hl, P.; and Misra, I. 2022{\natexlab{b}}.
\newblock Detecting Twenty-thousand Classes using Image-level Supervision.
\newblock In \emph{ECCV}.

\bibitem[{Zhu, Wang, and Saligrama(2020)}]{zhu2020don}
Zhu, P.; Wang, H.; and Saligrama, V. 2020.
\newblock Don't even look once: Synthesizing features for zero-shot detection.
\newblock In \emph{CVPR}.

\bibitem[{Zhu et~al.(2021)Zhu, Su, Lu, Li, Wang, and Dai}]{zhu2020deformable}
Zhu, X.; Su, W.; Lu, L.; Li, B.; Wang, X.; and Dai, J. 2021.
\newblock Deformable Detr: Deformable Transformers for End-To-End Object Detection.
\newblock In \emph{ICLR}.

\end{thebibliography}

\end{document}